\documentclass[10pt, a4paper]{article}
\pdfoutput=1
\usepackage{lrec}
\usepackage{multibib}
\newcites{languageresource}{Language Resources}
\usepackage{graphicx}
\usepackage{tabularx}
\usepackage{soul}

\usepackage{amsmath}
\usepackage{multirow}
\usepackage{enumitem}

\usepackage{epstopdf}
\usepackage[latin1]{inputenc}

\usepackage{hyperref}
\usepackage{xstring}

\newcommand{\watset}{\textsc{Watset}}
\newcommand{\watlink}{\textsc{Watlink}}
\newcommand{\subject}{Watasense}

\title{An Unsupervised Word Sense Disambiguation System for\\ Under-Resourced Languages}

\name{
  Dmitry Ustalov$^{\ast\dag}$, 
  Denis Teslenko$^\dag$,
  Alexander Panchenko$^\ddag$,
  Mikhail Chernoskutov$^\dag$, \\
  {\large\bf Chris Biemann$^\ddag$},
  {\large\bf Simone Paolo Ponzetto$^\ast$}
}

\address{
  $^\ast$ Data and Web Science Group, University of Mannheim, Germany \\
  $^\dag$ Ural Federal University, Russia \\
  $^\ddag$ Universit\"{a}t Hamburg, Department of Informatics, Language Technology Group, Germany \\
  \{dmitry,simone\}@informatik.uni-mannheim.de,
  teslenkoden@gmail.com,\\
  mikhail.chernoskutov@urfu.ru,
  \{panchenko,biemann\}@informatik.uni-hamburg.de
 }

\abstract{In this paper, we present {\subject}, an unsupervised system for word sense disambiguation. Given a sentence, the system chooses the most relevant sense of each input word with respect to the semantic similarity between the given sentence and the synset constituting the sense of the target word. {\subject} has two modes of operation. The sparse mode uses the traditional vector space model to estimate the most similar word sense corresponding to its context. The dense mode, instead, uses synset embeddings to cope with the sparsity problem. We describe the architecture of the present system and also conduct its evaluation on three different lexical semantic resources for Russian. We found that the dense mode substantially outperforms the sparse one on all datasets according to the adjusted Rand index.\\\newline}

\begin{document}

\maketitleabstract

\section{Introduction}

Word sense disambiguation (WSD) is a natural language processing task of identifying the particular word senses of polysemous words used in a sentence. Recently, a lot of attention was paid to the problem of WSD for the Russian language~\cite{Lopukhin:16,Lopukhin:17,Ustalov:17:acl}. This problem is especially difficult because of both linguistic issues -- namely, the rich morphology of Russian and other Slavic languages in general -- and technical challenges like the lack of software and language resources required for addressing the problem.

To address these issues, we present {\subject}, an unsupervised system for word sense disambiguation. We describe its architecture and conduct an evaluation on three datasets for Russian. The choice of an unsupervised system is motivated by the absence of resources that would enable a supervised system for under-resourced languages. {\subject} is not strictly tied to the Russian language and can be applied to any language for which a tokenizer, part-of-speech tagger, lemmatizer, and a sense inventory are available.

The rest of the paper is organized as follows. Section~2 reviews related work. Section~3 presents the {\subject} word sense disambiguation system, presents its architecture, and describes the unsupervised word sense disambiguation methods bundled with it. Section~4 evaluates the system on a gold standard for Russian. Section~5 concludes with final remarks.

\section{Related Work}\label{sec:related}

Although the problem of WSD has been addressed in many SemEval campaigns~\cite[\textit{inter alia}]{Navigli:07,Agirre:10,Manandhar:10}, we focus here on word sense disambiguation \textit{systems} rather than on the research methodologies.

Among the freely available systems, IMS (``It Makes Sense'') is a supervised WSD system designed initially for the English language~\cite{Zhong:10}. The system uses a support vector machine classifier to infer the particular sense of a word in the sentence given its contextual sentence-level features. Pywsd is an implementation of several popular WSD algorithms implemented in a library for the Python programming language.\footnote{https://github.com/alvations/pywsd} It offers both the classical Lesk algorithm for WSD and path-based algorithms that heavily use the WordNet and similar lexical ontologies. DKPro~WSD~\cite{Miller:13} is a general-purpose framework for WSD that uses a lexical ontology as the sense inventory and offers the variety of WordNet-based algorithms. Babelfy~\cite{Moro:14} is a WSD system that uses BabelNet, a large-scale multilingual lexical ontology available for most natural languages. Due to the broad coverage of BabelNet, Babelfy offers entity linking as part of the WSD functionality.

\newcite{Panchenko:17:emnlp} present an unsupervised WSD system that is also knowledge-free: its sense inventory is induced based on the JoBimText framework, and disambiguation is performed by computing the semantic similarity between the context and the candidate senses~\cite{Biemann:13}. \newcite{Pelevina:16} proposed a similar approach to WSD, but based on dense vector representations (word embeddings), called SenseGram. Similarly to SenseGram, our WSD system is based on averaging of word embeddings on the basis of an automatically induced sense inventory. A crucial difference, however, is that we induce our sense inventory from synonymy dictionaries and not distributional word vectors. While this requires more manually created resources, a potential advantage of our approach is that the resulting inventory contains less noise.

\begin{figure*}[t]
\centering
\includegraphics[width=\textwidth]{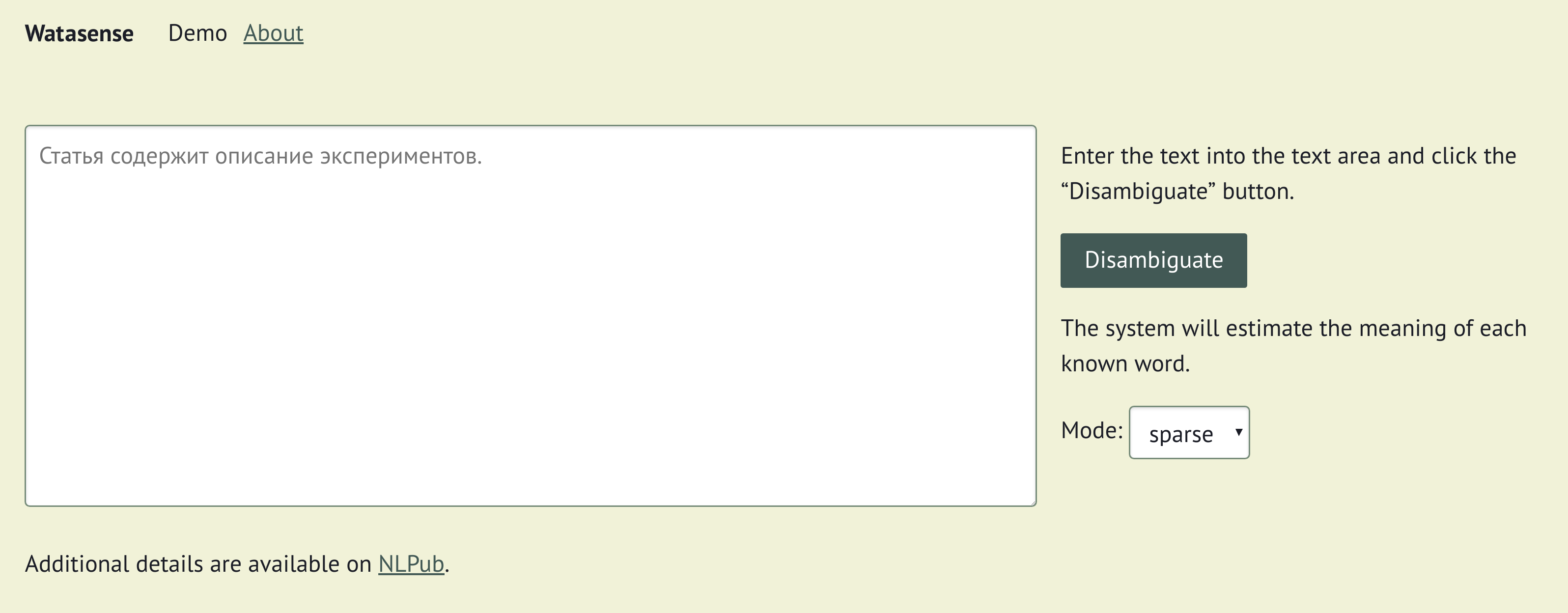}
\caption{A snapshot of the online demo, which is available at {http://watasense.nlpub.org/} (in Russian).}
\label{fig:input}
\end{figure*}

\section{{\subject}, an Unsupervised System for Word Sense Disambiguation}

{\subject} is implemented in the Python programming language using the scikit-learn~\cite{Pedregosa:11} and Gensim~\cite{Rehurek:10} libraries. {\subject} offers a Web interface (Figure~\ref{fig:input}), a command-line tool, and an application programming interface (API) for deployment within other applications.

\subsection{System Architecture}

A sentence is represented as a list of \textit{spans}. A span is a quadruple: $(w, p, l, i)$, where $w$ is the word or the token, $p$ is the part of speech tag, $l$ is the lemma, $i$ is the position of the word in the sentence. These data are provided by tokenizer, part-of-speech tagger, and lemmatizer that are specific for the given language. The WSD results are represented as a map of spans to the corresponding word sense identifiers. 

The sense inventory is a list of synsets. A synset is represented by three bag of words: the synonyms, the hypernyms, and the union of two former -- the \textit{bag}. Due to the performance reasons, on initialization, an inverted index is constructed to map a word to the set of synsets it is included into.

\begin{figure}[t]
\centering
\includegraphics[width=\columnwidth]{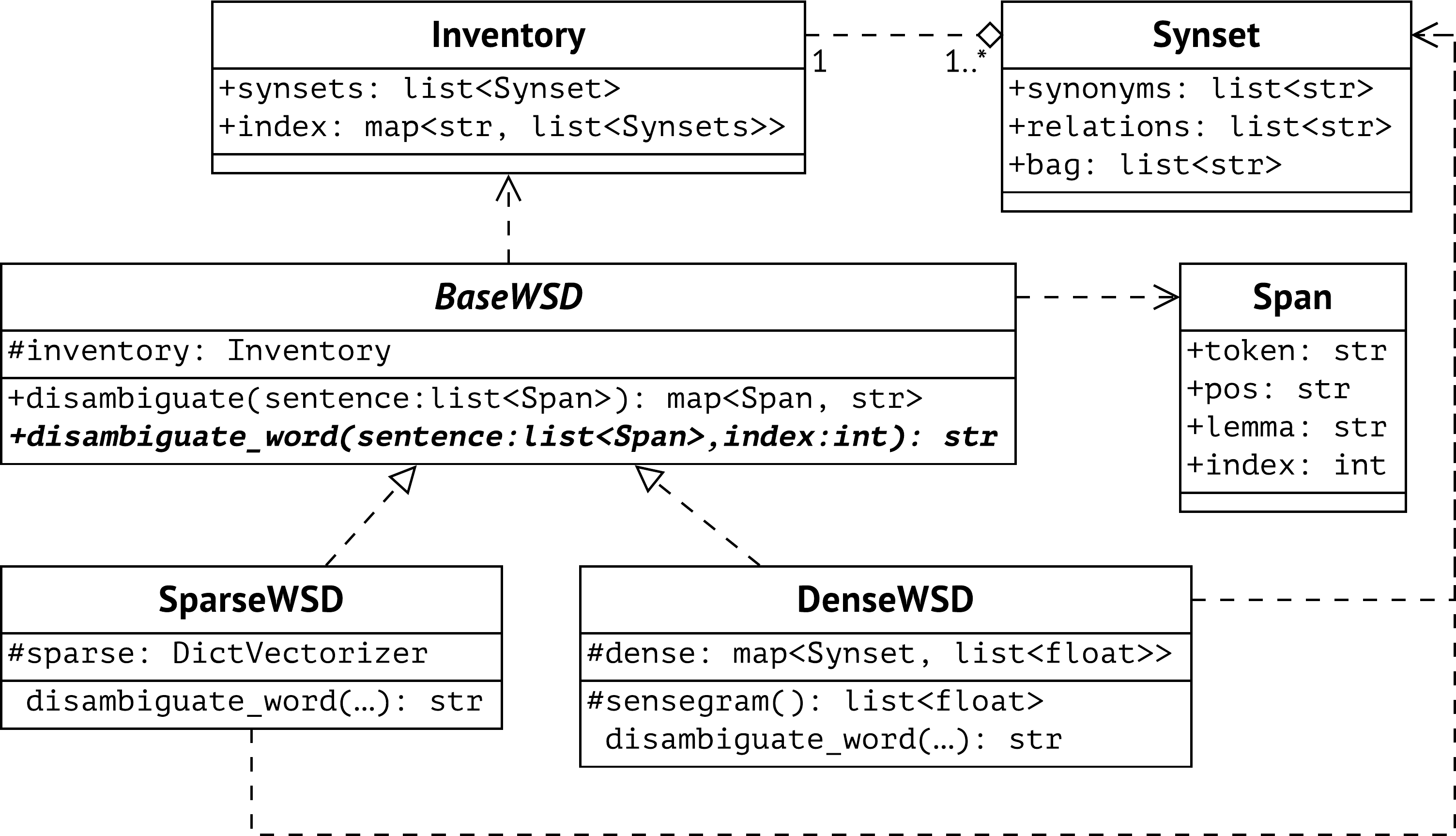}
\caption{The UML class diagram of {\subject}.}
\label{fig:uml-class}
\end{figure}

Each word sense disambiguation method extends the \texttt{BaseWSD} class. This class provides the end user with a generic interface for WSD and also encapsulates common routines for data pre-processing. The inherited classes like \texttt{SparseWSD} and \texttt{DenseWSD} should implement the \texttt{disambiguate\_word(\dots)} method that disambiguates the given word in the given sentence. Both classes use the \textit{bag} representation of synsets on the initialization. As the result, for WSD, not just the synonyms are used, but also the hypernyms corresponding to the synsets. The UML class diagram is presented in Figure~\ref{fig:uml-class}.

{\subject} supports two sources of word vectors: it can either read the word vector dataset in the binary Word2Vec format or use Word2Vec-Pyro4, a general-purpose word vector server.\footnote{https://github.com/nlpub/word2vec-pyro4} The use of a remote word vector server is recommended due to the reduction of memory footprint per each {\subject} process.

\subsection{User Interface}

\begin{figure*}[t]
\centering
\includegraphics[width=\textwidth]{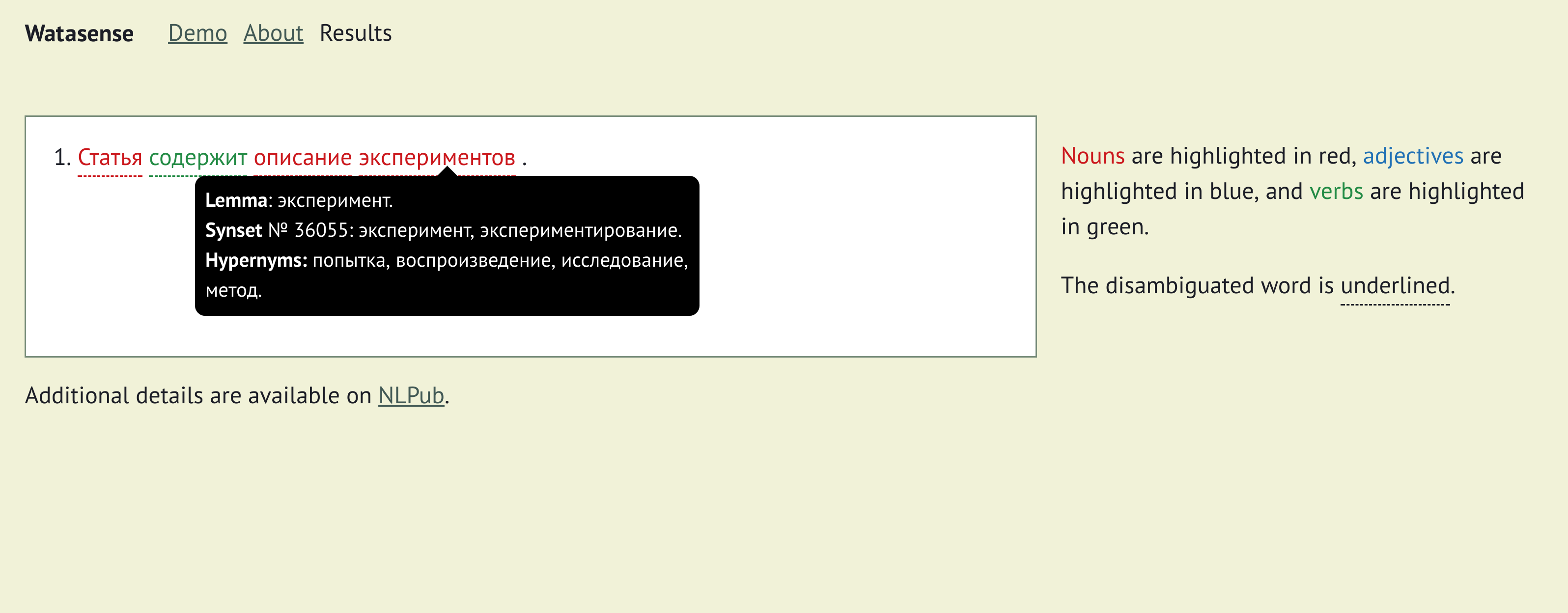}
\caption{The word sense disambiguation results with the word ``experiments'' selected. The tooltip shows its lemma ``experiment'', the synset identifier (36055), and the words forming the synset ``experiment'', ``experimenting'' as well as its hypernyms ``attempt'', ``reproduction'', ``research'', ``method''.}
\label{fig:result}
\end{figure*}

\figurename~\ref{fig:input} shows the Web interface of {\subject}. It is composed of two primary activities. The first is the text input and the method selection (\figurename~\ref{fig:input}). The second is the display of the disambiguation results with part of speech highlighting (\figurename~\ref{fig:result}). Those words with resolved polysemy are underlined; the tooltips with the details are raised on hover.

\subsection{Word Sense Disambiguation}

We use two different unsupervised approaches for word sense disambiguation. The first, called `sparse model', uses a straightforward sparse vector space model, as widely used in Information Retrieval, to represent contexts and synsets. The second, called `dense model', represents synsets and contexts in a dense, low-dimensional space by averaging word embeddings.

\paragraph{Sparse Model.} In the vector space model approach, we follow the sparse context-based disambiguated method~\cite{Faralli:16,Panchenko:17:emnlp}. For estimating the sense of the word $w$ in a sentence, we search for such a synset $\hat{w}$ that maximizes the cosine similarity to the sentence vector:
\begin{equation}
  \hat{w} = \arg\max_{S \ni w} \cos(S, T)\text{,}
\end{equation}
where $S$ is the set of words forming the synset, $T$ is the set of words forming the sentence. On initialization, the synsets represented in the sense inventory are transformed into the $\mathrm{tf}\textrm{--}\mathrm{idf}$-weighted word-synset sparse matrix efficiently represented in the memory using the compressed sparse row format. Given a sentence, a similar transformation is done to obtain the sparse vector representation of the sentence in the same space as the word-synset matrix. Then, for each word to disambiguate, we retrieve the synset containing this word that maximizes the cosine similarity between the sparse sentence vector and the sparse synset vector. Let $w_{\max}$ be the maximal number of synsets containing a word and $S_{\max}$ be the maximal size of a synset. Therefore, disambiguation of the whole sentence $T$ requires $O(|T| \times w_{\max} \times S_{\max})$ operations using the efficient sparse matrix representation.

\paragraph{Dense Model.} In the synset embeddings model approach, we follow SenseGram~\cite{Pelevina:16} and apply it to the synsets induced from a graph of synonyms. We transform every synset into its dense vector representation by averaging the word embeddings corresponding to each constituent word:
\begin{equation}
  \vec{S} = \frac{1}{|S|} \sum_{w \in S} \vec{w}\text{,}
\end{equation}
where $\vec{w}$ denotes the word embedding of $w$. We do the same transformation for the sentence vectors. Then, given a word $w$, a sentence $T$, we find the synset $\hat{w}$ that maximizes the cosine similarity to the sentence:
\begin{equation}
  \hat{w} = \arg\max_{S \ni w} \cos(\frac{\sum_{u \in S} \vec{u}}{|S|}, \frac{\sum_{u \in T} \vec{u}}{|T|})\text{.}
\end{equation}
On initialization, we pre-compute the dense synset vectors by averaging the corresponding word embeddings. Given a sentence, we similarly compute the dense sentence vector by averaging the vectors of the words belonging to non-auxiliary parts of speech, i.e., nouns, adjectives, adverbs, verbs, etc. Then, given a word to disambiguate, we retrieve the synset that maximizes the cosine similarity between the dense sentence vector and the dense synset vector. Thus, given the number of dimensions $d$, disambiguation of the whole sentence $T$ requires $(|T| \times w_{\max} \times d)$ operations.

\section{Evaluation}\label{sec:evaluation}

We conduct our experiments using the evaluation methodology of SemEval 2010 Task 14: Word Sense Induction \& Disambiguation~\cite{Manandhar:10}. In the gold standard, each word is provided with a set of instances, i.e., the sentences containing the word. Each instance is manually annotated with the single sense identifier according to a pre-defined sense inventory. Each participating system estimates the sense labels for these ambiguous words, which can be viewed as a clustering of instances, according to sense labels. The system's clustering is compared to the gold-standard clustering for evaluation.

\subsection{Quality Measure}

The original SemEval 2010 Task 14 used the V-Measure external clustering measure~\cite{Manandhar:10}. However, this measure is maximized by clustering each sentence into his own distinct cluster, i.e., a `dummy' singleton baseline. This is achieved by the system deciding that every ambiguous word in every sentence corresponds to a different word sense. To cope with this issue, we follow a similar study~\cite{Lopukhin:17} and use instead of the adjusted Rand index (ARI) proposed by \newcite{Hubert:85} as an evaluation measure.

In order to provide the overall value of ARI, we follow the addition approach used in~\cite{Lopukhin:17}. Since the quality measure is computed for each lemma individually, the total value is a weighted sum, namely
\begin{equation}
  \mathrm{ARI} = \frac{1}{\sum_w \left|I(w)\right|} \sum_w \mathrm{ARI}_w \times  \left|I(w)\right|\text{,}
  \label{eq:ari}
\end{equation}
where $w$ is the lemma, $I(w)$ is the set of the instances for the lemma $w$, $\mathrm{ARI}_w$ is the adjusted Rand index computed for the lemma $w$. Thus, the contribution of each lemma to the total score is proportional to the number of instances of this lemma.

\subsection{Dataset}

We evaluate the word sense disambiguation methods in {\subject} against three baselines: an unsupervised approach for learning multi-prototype word embeddings called AdaGram~\cite{Bartunov:16}, same sense for all the instances per lemma (One), and one sense per instance (Singletons). The AdaGram model is trained on the combination of RuWac, Lib.Ru, and the Russian Wikipedia with the overall vocabulary size of 2 billion tokens~\cite{Lopukhin:17}.

As the gold-standard dataset, we use the WSD training dataset for Russian created during RUSSE'2018: A Shared Task on Word Sense Induction and Disambiguation for the Russian Language~\cite{Panchenko:18:russe}. The dataset has $31$ words covered by $3\,491$ instances in the \textit{bts-rnc} subset and $5$ words covered by $439$ instances in the \textit{wiki-wiki} subset.\footnote{http://russe.nlpub.org/2018/wsi/}

The following different sense inventories have been used during the evaluation:
\begin{itemize}[leftmargin=4mm]
  \item \textbf{\watlink}, a word sense network constructed automatically. It uses the synsets induced in an unsupervised way by the \watset{[CW\textsubscript{nolog}, MCL]} method~\cite{Ustalov:17:acl} and the semantic relations from such dictionaries as Wiktionary referred as \textit{Joint$+$Exp$+$SWN} in~\newcite{Ustalov:17:dialogue}. This is the only automatically built inventory we use in the evaluation.
  \item \textbf{RuThes}, a large-scale lexical ontology for Russian created by a group of expert lexicographers~\cite{Loukachevitch:11}.\footnote{http://www.labinform.ru/pub/ruthes/index\_eng.htm}
  \item \textbf{RuWordNet}, a semi-automatic conversion of the RuThes lexical ontology into a WordNet-like structure~\cite{Loukachevitch:16}.\footnote{http://www.labinform.ru/pub/ruwordnet/index\_eng.htm}
\end{itemize}

Since the \textit{Dense} model requires word embeddings, we used the 500-dimensional word vectors from the Russian Distributional Thesaurus~\cite{Panchenko:17:aist}.\footnote{https://doi.org/10.5281/zenodo.400631} These vectors are obtained using the Skip-gram approach trained on the \texttt{lib.rus.ec} text corpus.

\subsection{Results}

We compare the evaluation results obtained for the \textit{Sparse} and \textit{Dense} approaches with three baselines: the AdaGram model (AdaGram), the same sense for all the instances per lemma (One) and one sense per instance (Singletons). The evaluation results are presented in Table~\ref{tab:results}. The columns bts-rnc and wiki-wiki represent the overall value of ARI according to Equation~\eqref{eq:ari}. The column Avg.\ consists of the weighted average of the datasets w.r.t.\ the number of instances.

\begin{table}[t]
\centering
\caption{Results on RUSSE'2018 (Adjusted Rand Index).}
\label{tab:results}
\begin{tabular}{|lc|cc|c|}
\hline
\multicolumn{2}{|l|}{\textbf{Method}} &
\textbf{bts-rnc} &
\textbf{wiki-wiki} &
\textbf{Avg.} \\\hline
\multicolumn{2}{|l|}{AdaGram}  & $0.22$
                              & $0.39$
                              & $0.23$ \\\hline
\multirow{2}{*}{{\watlink}}  & \textit{Sparse}
                             & $0.01$
                             & $0.07$
                             & $0.01$ \\
                             & \textit{Dense}
                             & $0.08$
                             & $0.14$
                             & $0.08$ \\\hline
\multirow{2}{*}{RuThes}      & \textit{Sparse}
                             & $0.00$
                             & $0.17$
                             & $0.01$ \\
                             & \textit{Dense}
                             & $0.14$
                             & $0.47$
                             & $0.17$ \\\hline
\multirow{2}{*}{RuWordNet}   & \textit{Sparse}
                             & $0.00$
                             & $0.11$
                             & $0.01$ \\
                             & \textit{Dense}
                             & $0.12$
                             & $0.50$
                             & $0.15$ \\\hline
\multicolumn{2}{|l|}{One}    & $0.00$
                             & $0.00$
                             & $0.00$ \\
\multicolumn{2}{|l|}{Singletons} & $0.00$
                                 & $0.00$
                                 & $0.00$ \\
\hline
\end{tabular}
\vspace{-1em}
\end{table}

We observe that the SenseGram-based approach for word sense disambiguation yields substantially better results in every case (Table~\ref{tab:results}). The primary reason for that is the implicit handling of similar words due to the averaging of dense word vectors for semantically related words. Thus, we recommend using the dense approach in further studies. Although the AdaGram approach trained on a large text corpus showed better results according to the weighted average, this result does not transfer to languages with less available corpus size.

\section{Conclusion}\label{sec:conclusion}

In this paper, we presented {\subject},\footnote{https://github.com/nlpub/watasense} an open source unsupervised word sense disambiguation system that is parameterized only by a word sense inventory. It supports both sparse and dense sense representations. We were able to show that the dense approach substantially boosts the performance of the sparse approach on three different sense inventories for Russian. We recommend using the dense approach in further studies due to its smoothing capabilities that reduce sparseness. In further studies, we will look at the problem of phrase neighbors that influence the sentence vector representations.

Finally, we would like to emphasize the fact that {\subject} has a simple API for integrating different algorithms for WSD. At the same time, it requires only a basic set of language processing tools to be available: tokenizer, a part-of-speech tagger, lemmatizer, and a sense inventory, which means that low-resourced language can benefit of its usage.

\section{Acknowledgements}

We acknowledge the support of the Deutsche Forschungsgemeinschaft (DFG) under the project ``Joining Ontologies and Semantics Induced from Text'' (JOIN-T), the RFBR under the projects no.~16-37-00203~mol\_a and no.~16-37-00354~mol\_a, and the RFH under the project no.~16-04-12019. The research was supported by the Ministry of Education and Science of the Russian Federation Agreement no.~02.A03.21.0006. The calculations were carried out using the supercomputer ``Uran'' at the Krasovskii Institute of Mathematics and Mechanics.

\section{Bibliographical References}

\bibliographystyle{lrec}
\bibliography{wsd.lrec2018}



\end{document}